\documentclass[conference, A4]{IEEEtran}
\IEEEoverridecommandlockouts
\usepackage{cite}
\usepackage{amsmath,amssymb,amsfonts}
\usepackage{algorithmic}
\usepackage{graphicx}
\usepackage{textcomp}
\usepackage{xcolor}

\usepackage{subfig}
\usepackage{multirow}

\def\BibTeX{{\rm B\kern-.05em{\sc i\kern-.025em b}\kern-.08em
    T\kern-.1667em\lower.7ex\hbox{E}\kern-.125emX}}
\begin{document}

\title{Neural Authorship Attribution: Stylometric Analysis on Large Language Models}

\author{\IEEEauthorblockN{Tharindu Kumarage}
\IEEEauthorblockA{\textit{School of Computing and Augmented Intelligence} \\
\textit{Arizona State University}\\
kskumara@asu.edu}
\and
\IEEEauthorblockN{Huan Liu}
\IEEEauthorblockA{\textit{School of Computing and Augmented Intelligence} \\
\textit{Arizona State University}\\
% City, Country \\
huanliu@asu.edu}
}

\maketitle

\begin{abstract}

Large language models (LLMs) such as GPT-4, PaLM, and Llama have significantly propelled the generation of AI-crafted text. With rising concerns about their potential misuse, there is a pressing need for AI-generated-text forensics. Neural authorship attribution is a forensic effort, seeking to trace AI-generated text back to its originating LLM. The LLM landscape can be divided into two primary categories: proprietary and open-source. In this work, we delve into these emerging categories of LLMs, focusing on the nuances of neural authorship attribution. To enrich our understanding, we carry out an empirical analysis of LLM writing signatures, highlighting the contrasts between proprietary and open-source models, and scrutinizing variations within each group. By integrating stylometric features across lexical, syntactic, and structural aspects of language, we explore their potential to yield interpretable results and augment pre-trained language model-based classifiers utilized in neural authorship attribution. Our findings, based on a range of state-of-the-art LLMs, provide empirical insights into neural authorship attribution, paving the way for future investigations aimed at mitigating the threats posed by AI-generated misinformation.

\end{abstract}


\section{Introduction}

Recent advancements in generative large language models (LLMs), e.g., proprietary models such as OpenAI's GPT-4 \cite{openai2023gpt4}, and Google's PaLM \cite{chowdhery2022palm}, along with open-source models like Llama 1\&2 \cite{touvron2023llama} have led to an unparalleled proliferation of AI-generated content, predominantly in text form. Although AI-generated text can significantly enhance human productivity across various writing tasks, there are substantial risks, as adversaries could exploit these LLMs to conduct influence operations or disseminate false information, thereby threatening cybersecurity and the integrity of the information ecosystem. For instance, a surge in AI-generated news articles, often appearing to be authentic but containing deceptive content, is particularly concerning \cite{goldstein2023can, subbiah2023towards}. Consequently, there is an urgent need for computational methods that can support the forensic analysis of AI-generated-text to combat such LLM-propelled misinformation campaigns.

A pivotal component of AI-generated-text forensics involves identifying the source LLM used to generate a specific text, a process referred to as neural authorship attribution %due to the authorship being attributed to LLMs 
~\cite{uchendu2023attribution}. The importance of neural authorship attribution cannot be overstated, as it provides a critical function within the broader forensic process. The ability to attribute a piece of AI-generated text to a particular LLM aids in unmasking the underlying characteristics of malicious actors and campaigns that use these LLMs. It provides key insights necessary for crafting appropriate countermeasures, and helps enhance the security protocols or refining the ethical guidelines associated with the usage of LLMs. %The field of neural authorship attribution has seen considerable advancements through numerous research efforts. 
The methods of neural authorship attribution typically involve training a classifier on the embeddings produced by the pre-trained language models (PLMs), such as RoBERTa for a corpus of texts, where each text is generated by a known LLM. The classifier then learns the nuanced differences in the text generation styles of different LLMs, thereby equipping it to attribute new texts to the correct LLM. However, the rapid evolution of the LLM landscape has given rise to new dimensions in neural authorship attribution. 

One such novel dimension pertains to the categorization of the source LLM: proprietary and open-source. Proprietary LLMs are owned by a specific organization, and accessible through a paywall via an API; open-source LLMs are freely available models that can be run on one's own computing infrastructure. We suggest that determining whether the source LLM is proprietary or open-source holds substantial merit for forensics. This step could provide additional characteristics of the misinformation campaign, such as the financial resources and level of expertise of the orchestrating malicious actors. For example, a campaign using an open-source LLM rather than a proprietary one may indicate the presence of specialized expertise and a dedicated computational infrastructure necessary to operate these models.

In this paper, we aim to enhance our understanding of this newly emerging dimension of neural authorship attribution. This is achieved by empirically studying the writing signatures of LLMs in two stages: 1) comparing proprietary LLMs with open-source LLMs to identify commonalities or disparities in their writing signatures, and 2) investigating the variability of writing signatures within each sub-classification among state-of-the-art LLMs. To facilitate the interpretability of our empirical analysis, we measure the writing signatures of LLMs by incorporating comprehensive stylometric features that span across lexical, syntactic, and structural dimensions of natural language. Subsequently, we conduct experiments exploring how these interpretable stylometric features can be combined with powerful PLM-based classifiers to improve neural authorship attribution. Our findings paves the way for further research in neural authorship attribution.

\section{Related Work}

% \subsection{Authorship Attribution}

Authorship Attribution (AA), a task focused on identifying authors by their distinct writing signatures, has been widely explored. In its early stages, classical classifiers like Naive Bayes, SVM, Conditional Tree, Random Forest, and KNN ~\cite{koppel2009computational} were employed in conjunction with feature extraction techniques such as n-grams, POS-tags, topic modeling, and LIWC to solve AA problems~\cite{uchendu2023attribution}. However, with the advancements in neural networks, models like Convolutional Neural Networks (CNNs) ~\cite{boumber-etal-2018-experiments} and Recurrent Neural Networks (RNNs) ~\cite{alsulami2017source} have emerged as more appropriate for AA, given their ability to represent an author's unique characteristics accurately. 

With transformer-based language models, a unique category of authors, known as neural authors, has been incorporated into AA task landscape~\cite{uchendu2023attribution}. The task here involves identifying the source language model that generated a specific text. In the early stages, the same stylometric and statistical features used in traditional AA were applied to neural authors to facilitate a better understanding of the task \cite{uchendu-etal-2020-authorship, kumarage2023stylometric}. More recently, studies have incorporated pre-trained language model (PLM) based classifiers to determine the authorship of neural authors, such as GPT-2, GPT-3, and Grover ~\cite{frohling2021feature}. The most recent related work has involved efforts to attribute the authorship of fine-tuned instances to the base LLM ~\cite{foley-etal-2023-matching}.

\section{Methodology}

In this study, we aim to quantify the writing signatures of various LLMs (proprietary or open-source), enabling more interpretable neural authorship attribution. Our methodology comprises three steps. First, we compiled a dataset incorporating texts from six state-of-the-art LLMs, representative of the two categories under examination. Next, we extracted the writing signatures of these LLMs, integrating 60 stylometric features spanning lexical, syntactic, and structural language dimensions. Finally, we incorporated these features into our interpretable neural authorship attribution model.

\subsection{Dataset Generation}
In this study, we examined six LLMs: GPT-4, GPT-3.5, Llama 1, Llama 2, and GPT-NeoX. GPT-4 and GPT-3.5 are proprietary models, whereas the remaining are open-source. Our analysis was confined to news articles generated by these LLMs to control for domain-specific variations in the resulting text, which can affect distinct writing characteristics. We adopted the prompting technique from previous studies, where the headline of a pristine human-authored article was used to prompt the LLMs to generate a corresponding AI-generated news article \cite{uchendu2023attribution}. We sampled 1K headlines from the dataset provided in \cite{uchendu2023attribution}, encompassing news articles from CNN and the Washington Post. Subsequently, using these 1K headlines, we prompted the six chosen LLMs to generate corresponding news articles. This process yielded a dataset of 6K AI-generated news articles, with each LLM contributing 1K instances.

\noindent \textbf{Proprietary LLMs:} Data from GPT-4 and GPT-3.5 were generated using OpenAI's API. To support the reproducibility of our work, we selected snapshot versions \textit{'gpt-4-0613'} and \textit{'gpt-3.5-turbo-0613'} from the API.

\noindent \textbf{Open-Source LLMs:} We selected Llama models (7B parameters variants), as they currently rank highly on the Open LLM Leaderboard hosted on Hugging Face, under the pre-trained model category\footnote{https://huggingface.co/spaces/HuggingFaceH4/open\_llm\_leaderboard}. We included GPT-NeoX (20B) in our study due to its prominent rankings in the past, enabling us to assess the progression of writing signatures in open-source LLMs.

\noindent \textbf{Generation Configurations:} The generation configurations applied when generating text through these LLMs warrant attention, as different configurations on the same model can lead to distinct styles and text diversity. In this study, we set $top\_p$ to 0.9 and $temperature$ to 0.8.

\subsection{Extraction of the Writing Signature} 
We assess each model's writing signature by applying a set of stylometric features. Specifically, for each AI-generated news article within our dataset, we extract stylometric features along the axes of lexical, syntactic, and structural attributes of the text. For convenience, we will denote the stylometric feature set as $F$. In this context, $F$ can be divided into three subsets, $F=F^1 \cup F^2 \cup F^3$, where $F^i| i \in {1, 2, 3}$ respectively represent lexical, syntactic, and structural features. These subsets can be broadly defined as follows:
\begin{enumerate} 
	\item $F^1$: \textbf{Lexical Features} - They quantify the unique style of an author's word usage. Our analysis includes measurements of average word length, function word usage, unique word richness using a moving average type-token ratio (MTTR) for vocabulary richness, hapax legomena (words that appear only once), and stopword usage. 
	\item $F^2$: \textbf{Syntactic Features} - These features quantify the author's use of sentence grammatical structure. Our analysis measures average sentence length (number of words per sentence), usage of different parts of speech (POS) (frequency of nouns, verbs, adjectives, adverbs, pronouns, etc., per sentence), as well as usage of active vs. passive voice and past vs. present tense. 
	\item $F^3$: \textbf{Structural Features} - They quantify the organization and layout of the text. In this category, we measure average paragraph length (word and sentence frequency per paragraph), frequency of punctuation usage (!, ', ,, :, ;, ?, ", -,--, @, \#), and the usage of capital letters.
\end{enumerate}

For the features where we compute a mean score such as average word length, we also calculate the standard deviation (std) and add that as another stylometry feature. Moreover, certain features in our study yield values greater than 1 such as stopword usage, while some features are mean and std values within the range of [0,1]. Thus, before incorporating the final writing signature vector into our analysis and the task of neural authorship attribution, we normalize it. Let $n$ represent the number of stylometric features, $f_i \in F$, and $W =( w_{1}, w_{2},\dots, w_{m})$ denote the treebank tokenization \footnote{https://www.nltk.org/api/nltk.tokenize.word\_tokenize.html} of the input AI-generated news article. We then define the final normalized writing signature vector $S_{n}$ as follows:
\begin{equation} 
	S_{n} = \frac{[f_1(W), f_2(W), ..., f_n(W)]}{||[f_1(W), f_2(W), ..., f_n(W)]||}. 
	\label{eq:feat_extrc} 
\end{equation}

\subsection{Neural Authorship Attribution}

In alignment with existing work on neural author attribution, we treat this task as a classification problem~\cite{uchendu2023attribution, foley-etal-2023-matching}. During training, a classifier learns to delineate a decision boundary within the writing signature space. This allows the classifier to attribute a given text to one of the known LLMs in the training data during testing. However, our approach divides this classification task into several sub-tasks. Initially, we conduct a binary classification task, training a classifier to distinguish whether a given text originates from a \textit{proprietary LLM} or an \textit{open-source LLM}. Subsequently, we perform multiple classification tasks to analyze neural authorship attribution within the proprietary and open-source categories. 

We balance our 6K data samples across the classes for each classification task. For instance, in the \textit{proprietary LLM} vs. \textit{open-source LLM} classification, we compile a dataset of 2K (with 9:1 train, test split) proprietary LLM articles by combining data from GPT-3.5 and GPT-4. Similarly, we create a 2K-article dataset from the open-source LLMs, assembling $\approx$0.66K samples from each of the three open-source LLMs. For our analysis, we utilize the following classification models:

\noindent \textbf{Classical Classification: } We incorporate XGBoost as the classifier. We use the stylometric writing signature features we extracted (XGB$_{stylo}$) and Bag-of-words features (XGB$_{bow}$). The motivation is to provide interpretable results by analyzing important stylometry features for neural authorship attribution. 

\noindent \textbf{PLM-based Classification: } We use the state-of-the-art PLM recently utilized for the neural authorship attribution task, namely, the RoBERTa model~\cite{uchendu2023attribution}. We deploy two variations of this model: one is the base RoBERTa without fine-tuning (RoBERTa$_{Zero}$), and the other is the RoBERTa model fine-tuned on our prepared classification datasets (RoBERTa$_{FT}$).

\noindent \textbf{PLM Stylometry Fusion: } We propose an approach where stylometry features synergize with the RoBERTa embeddings, which provide high semantic volume (RoBERTa$_{Stylo}$). Specifically, we form a concatenated vector that combines both representations. This fusion is accomplished through an attention layer, and the resulting combined representation is subsequently passed to the classifier.

\section{Results}

\subsection{Preliminary Analysis}

\begin{figure}[t]
    \centering
    \subfloat[PLM-embedding]{%
        \includegraphics[width=0.45\linewidth]{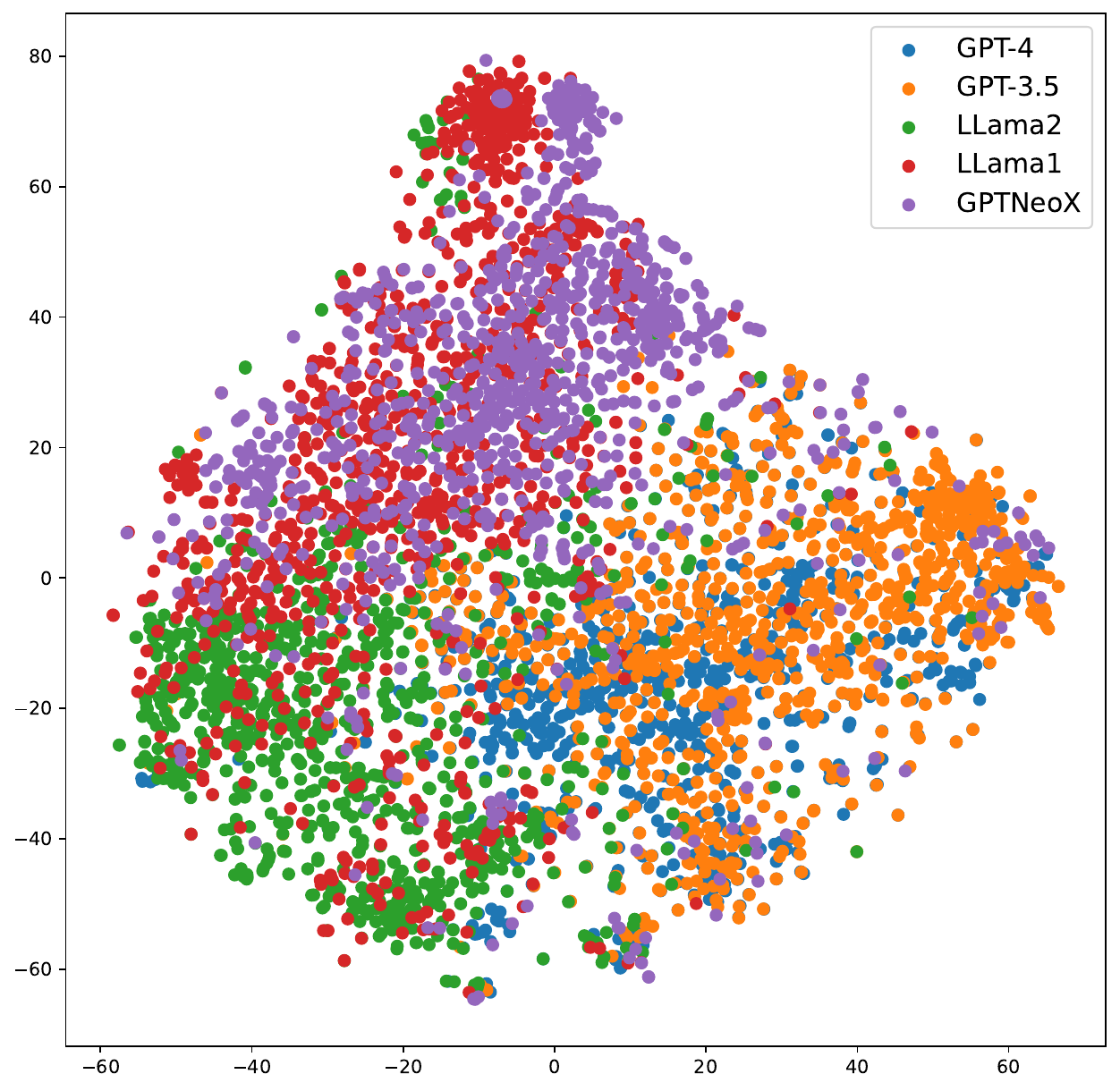}%
        \label{fig:tsne_plm}%
    }
    \subfloat[Stylometry-features]{%
        \includegraphics[width=0.45\linewidth]{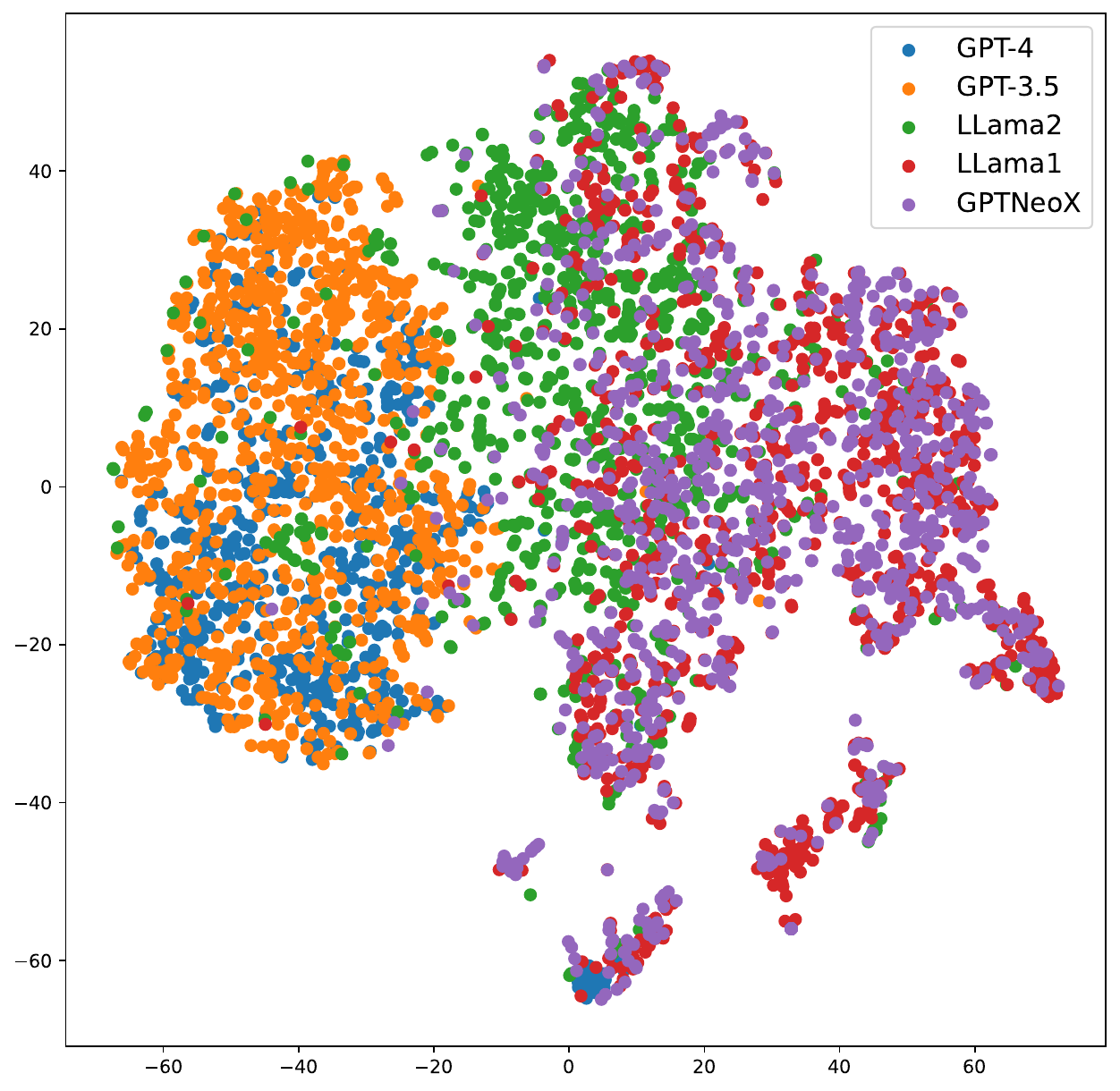}%
        \label{fig:tsne_stylo}%
    }
    \caption{t-SNE plots showcasing the distributions of different LLMs. (a) embedding space of RoBERTa-PLM and (b) stylometry feature space used to measure the writing signature.}
    \label{fig:tsne_plots}
\end{figure}

To gain deeper insight into the distinctiveness or similarities of the LLMs examined in our study, we carried out a preliminary analysis using t-SNE plots to visualize the distributions of each LLM. We applied t-SNE to two separate feature spaces: the RoBERTa embedding space and our stylometry feature space. The RoBERTa embedding space comprises 768 dimensions, while the stylometry space contains 60 dimensions. We employed t-SNE to derive a 2-dimensional representation from these two feature spaces. As observed in Figure \ref{fig:tsne_plots}, both feature spaces demonstrate a clear separation between \textit{proprietary} and \textit{open-source} LLMs. However, notably in Figure \ref{fig:tsne_stylo}, the stylometry space presents a more pronounced distinction between the two LLM categories, suggesting that stylometry features serve as robust signals beneficial for neural authorship attribution. Within each LLM category, there is a noticeable overlap in the distributions. For instance, there's a marked overlap between GPT-3.5 and GPT-4 data points within the proprietary LLMs. Similarly, in the open-source category, Llama 1 and GPT-NeoX exhibit considerable overlap. Intriguingly, Llama 2 data points tend to align more closely with proprietary LLMs such as GPT-4 and GPT-3.5 in the stylometry space, hinting that Llama 2 shares a similar writing style to those models. Furthermore, given that Llama 2 is reputed for its superior capability in generating engaging text compared to other existing open-source LLMs, this might indicate a narrowing gap between open-source LLMs and their more powerful proprietary counterparts.

\subsection{Proprietary vs. Open-source Attribution}

Table \ref{tab:results} presents the results for neural authorship attribution. In the table, the first column, labeled `Initial Attribution', displays outcomes for attributing a specific text to either the \textit{Proprietary LLM} or \textit{Open-source LLM} class. Our preliminary analysis, which indicated a marked differentiation between text of these two classes, is further validated by these results. Almost all classifiers demonstrate commendable performance in this initial attribution task, with our fusion approach achieving near-perfect scores. However, in a supplementary experiment that only considers Llama 2 data to represent the \textit{open-source} class, we saw an average performance decline of 7.4\%. This suggests that as open-source models advance, the initial stage of attribution becomes increasingly challenging.

\begin{figure} 
	\centering 
	\includegraphics[width=0.35\textwidth]{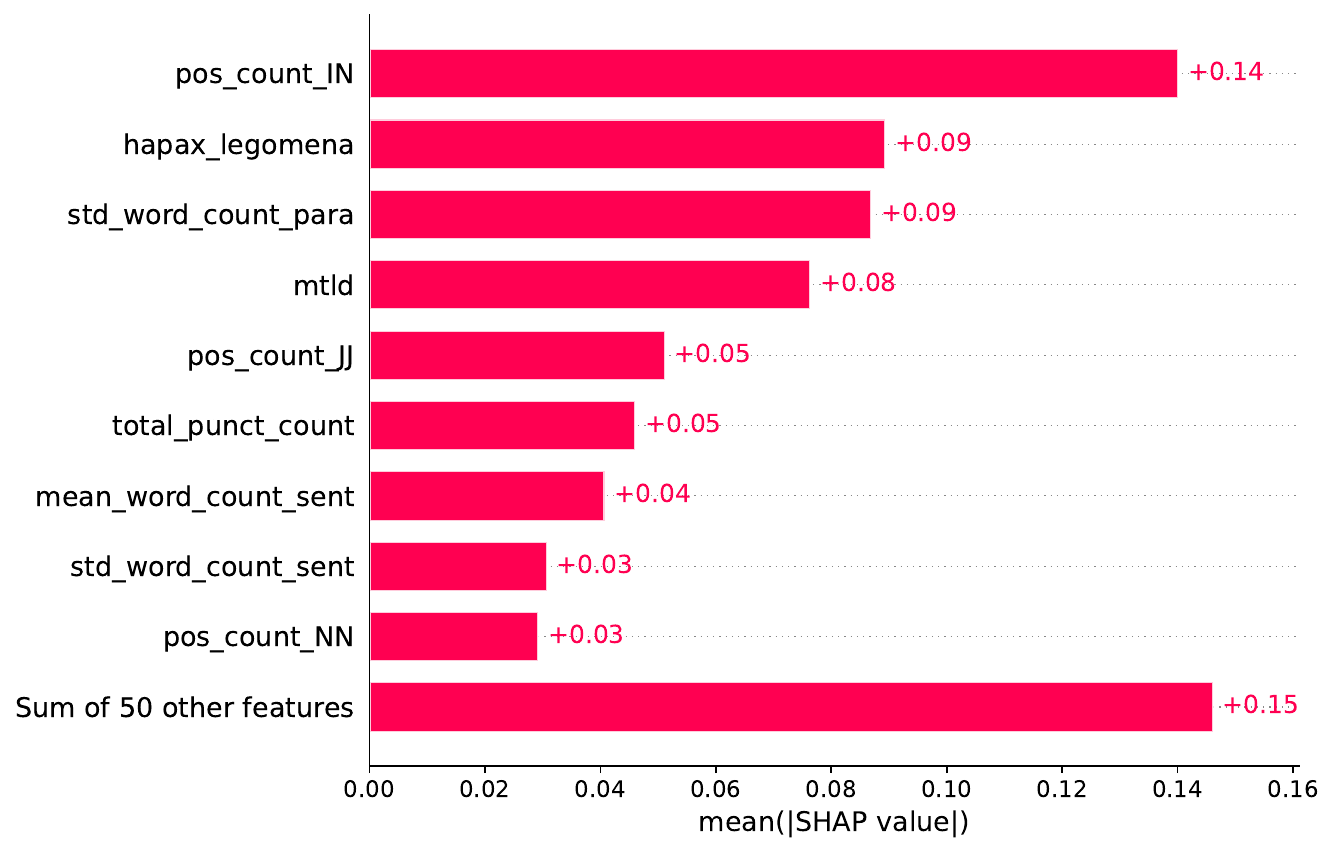} 	
	\caption{SHAP values showcasing feature importance in proprietary LLM vs. open-source LLM attribution.}
	\label{fig:feat_impo} 
\end{figure}

Furthermore, we conducted a study to elucidate the relevance of various stylometry features in the LLM writing signatures to interpret the \textit{proprietary} vs. \textit{open-source} attribution performances. We employed the Shapley Additive Explanations (SHAP) \cite{lundberg2017unified} on the XGBoost classifier to assess feature importance. As illustrated in Figure \ref{fig:feat_impo}, the SHAP plots underscore the vital role of lexical diversity in distinguishing the two LLM categories. This distinction is notably shown with features such as MTTR and hapax legomena, which command higher importance scores. Additionally, certain syntactic features, notably the usage of distinct parts of speech—including Prepositions (\textit{pos\_count\_IN}), Adjectives (\textit{pos\_count\_JJ}), and Nouns (\textit{pos\_count\_NN})—stand out in their significance. Moreover, structural attributes like punctuation usage and paragraph length are crucial in differentiating \textit{proprietary} from \textit{open-source} LLMs. Delving deeper into why these features manifest differently in various LLMs is a promising avenue for future research, which would further augment neural authorship attribution.

% \begin{table}[]
% \resizebox{\columnwidth}{!}{%
% \begin{tabular}{ccccccccc}
% \hline
% \multirow{2}{*}{\textbf{Model}} &
%   \multicolumn{2}{c}{\textbf{Initial Attribution}} &
%   \multicolumn{2}{c}{\textbf{Proprietary}} &
%   \multicolumn{3}{c}{\textbf{Open-Source}} \\ 
%   \cline{2-3} 
%   \cline{5-6} 
%   \cline{6-8} 
%  &
%   Prop. &
%   Op.So. &
%   \begin{tabular}[c]{@{}c@{}}GPT\\ 3.5\end{tabular} &
%   \begin{tabular}[c]{@{}c@{}}GPT\\ 4\end{tabular} &
%   \begin{tabular}[c]{@{}c@{}}GPT-\\ NeoX\end{tabular} &
%   Llama1 &
%   Llama2 \\ \hline
% XGB$_{BOW}$      & 0.858 & 0.845 & 0.742 & 0.747 & 0.305 & 0.346 & 0.539 \\
% XGB$_{Stylo}$    & 0.954 & 0.949 & 0.895 & 0.901 & 0.743 & 0.714 & 0.860 \\
% RoBERTa$_{Zero}$ & 0.881 & 0.873 & 0.809 & 0.816 & 0.653 & 0.672 & 0.819 \\
% RoBERTa$_{FT}$   & 0.972 & 0.968 & 0.925 & 0.929 & 0.754 & 0.725 & 0.898 \\
% RoBERTa$_{Stylo}$ &
%   \textbf{0.992} &
%   \textbf{0.987} &
%   \textbf{0.947} &
%   \textbf{0.951} &
%   \textbf{0.813} &
%   \textbf{0.802} &
%   \textbf{0.917} \\ \hline
% \end{tabular}%
% }
% \end{table}

\begin{table}[]
\resizebox{\columnwidth}{!}{%
\begin{tabular}{cccl@{}ccl@{}ccc}
\hline
\multirow{2}{*}{\textbf{Model}} &
  \multicolumn{2}{c}{\textbf{Initial Attribution}} &
   &
  \multicolumn{2}{c}{\textbf{Proprietary}} &
   &
  \multicolumn{3}{c}{\textbf{Open-Source}} \\ \cline{2-3} \cline{5-6} \cline{8-10} 
 &
  Prop. &
  Op.So. &
   &
  \begin{tabular}[c]{@{}c@{}}GPT\\ 3.5\end{tabular} &
  \begin{tabular}[c]{@{}c@{}}GPT\\ 4\end{tabular} &
   &
  \begin{tabular}[c]{@{}c@{}}GPT-\\ NeoX\end{tabular} &
  Llama1 &
  Llama2 \\ \hline
XGB$_{BOW}$      & 0.858 & 0.845 &  & 0.742 & 0.747 &  & 0.305 & 0.346 & 0.539 \\
XGB$_{Stylo}$    & 0.954 & 0.949 &  & 0.895 & 0.901 &  & 0.743 & 0.714 & 0.860 \\
RoBERTa$_{Zero}$ & 0.881 & 0.873 &  & 0.809 & 0.816 &  & 0.653 & 0.672 & 0.819 \\
RoBERTa$_{FT}$   & 0.972 & 0.968 &  & 0.925 & 0.929 &  & 0.754 & 0.725 & 0.898 \\
RoBERTa$_{Stylo}$ &
  \textbf{0.992} &
  \textbf{0.987} &
   &
  \textbf{0.947} &
  \textbf{0.951} &
   &
  \textbf{0.813} &
  \textbf{0.802} &
  \textbf{0.917} \\ \hline
\end{tabular}%
}
\caption{F-scores on three authorship attribution tasks: Initial attribution (proprietary vs. open-source),  Attribution within proprietary LLMs, and Attribution within open-source LLMs.}
\label{tab:results}
\end{table}

\subsection{Intra-category Attribution}

In Table \ref{tab:results}, the column labeled `Proprietary' presents performance scores for attributing texts to the proprietary LLMs GPT-4 and GPT-3.5. We observe that models augmented with stylometry demonstrate commendable accuracy in attributing texts to GPT-4 and GPT-3.5. This suggests that these two models exhibit unique writing signatures, despite originating from the same organization and having comparable underlying pretraining datasets. A probable explanation for this disparity might be the RLHF training steps \footnote{https://openai.com/research/instruction-following} applied to these models.
Similarly, we analyzed the attribution of texts to three open-source LLMs, as shown in Table \ref{tab:results} under the `Open-source' column. There is a significant decline in attribution performance for this task in comparison to the prior two tasks. It appears that GPT-NeoX and Llama 1 possess overlapping writing styles, resulting in diminished performance. Validating our observations from the preliminary analysis, Llama 2 exhibits notably higher attribution scores relative to the other two open-source models. This underscores Llama 2's distinct writing signature when contrasted with its predecessor open-source models.

\section{Conclusion}

In this study, we explored neural authorship attribution, diving deep into the nuances that differentiate proprietary and open-source LLMs. Through detailed stylometric analysis, we identified distinct differences in writing style between the two types, with lexical diversity, specific parts of speech, and structural features being key indicators. These insights not only pave the way for enhancing neural authorship attribution techniques but also have the potential to shed light on the evolution of these LLMs. The fact that GPT-4 and GPT-3.5, despite originating from the same organization, manifest distinct writing styles suggests that the training steps, like the RLHF, might play a pivotal role in shaping their outputs. The difficulty in distinguishing between individual open-source models hint at the convergence of their writing styles, possibly due to shared pre-training datasets or architecture similarities. However, Llama 2's distinct style shows potential for open-source models to rival proprietary counterparts. As we progress in this age of AI-generated content, further understanding these nuances of LLM outputs will be paramount, especially in the wake of potential misinformation threats.

% Entries for the entire Anthology, followed by custom entries
\bibliography{references}
\bibliographystyle{IEEEtran}

\end{document}